\documentclass{article} % For LaTeX2e
\usepackage{iclr2021_conference,times}

\usepackage{amsmath,amsfonts,bm}
\usepackage{amsthm}

\newcommand{\sald}{SALD}

\newcommand{\sal}{SAL}

\newcommand{\norm}[1]{\left\Vert#1\right\Vert}

\newcommand{\abs}[1]{\left\vert#1\right\vert}

\newcommand{\set}[1]{\left\{#1\right\}}
\newcommand{\parr}[1]{\left (#1\right )}
\newcommand{\brac}[1]{\left [#1\right ]}
\newcommand{\ip}[1]{\left \langle #1 \right \rangle }
\newcommand{\Real}{\mathbb R}

\newcommand{\too}{\rightarrow}

\newcommand{\diag}{\textrm{diag}} %diagonal matrix
\newcommand{\one}{\mathbf{1}}

\newcommand{\loss}{\mathrm{loss}}%_{\text{ur}}}
\newcommand{\eg}{{e.g.}}
\newcommand{\ie}{{i.e.}}

\makeatletter
\newtheorem*{rep@theorem}{\rep@title}
\newcommand{\newreptheorem}[2]{%
\newenvironment{rep#1}[1]{%
 \def\rep@title{#2 \ref{##1}}%
 \begin{rep@theorem}}%
 {\end{rep@theorem}}}
\makeatother

\newtheorem{theorem}{Theorem}
\newreptheorem{theorem}{Theorem}
\newtheorem{lemma}{Lemma}
\newreptheorem{lemma}{Lemma}

\usepackage{amsmath,amsfonts,bm}

\def\eqref#1{equation~\ref{#1}}
\def\1{\bm{1}}

\newcommand{\test}{\mathcal{D_{\mathrm{test}}}}

\def\vmu{{\bm{\mu}}}

\def\vtheta{{\bm{\theta}}}
\def\veta{{\bm{\eta}}}

\def\va{{\bm{a}}}
\def\vb{{\bm{b}}}

\def\vn{{\bm{n}}}

\def\vu{{\bm{u}}}

\def\vx{{\bm{x}}}

\def\vy{{\bm{y}}}
\def\vz{{\bm{z}}}

\def\mW{{\bm{W}}}
\def\mX{{\bm{X}}}
\def\mY{{\bm{Y}}}

\DeclareMathAlphabet{\mathsfit}{\encodingdefault}{\sfdefault}{m}{sl}
\SetMathAlphabet{\mathsfit}{bold}{\encodingdefault}{\sfdefault}{bx}{n}

\def\gD{{\mathcal{D}}}

\def\gN{{\mathcal{N}}}

\def\gS{{\mathcal{S}}}
\def\gT{{\mathcal{T}}}
\def\gU{{\mathcal{U}}}

\def\gX{{\mathcal{X}}}

\newcommand{\dist}{\textrm{d}} %distance function
\newcommand{\E}{\mathbb{E}}

\usepackage{hyperref}
\usepackage{url}
\usepackage{graphicx}
\usepackage{enumitem}
\usepackage{amsmath,amssymb} % define this before the line numbering.
\usepackage{amsthm}
\usepackage{wrapfig}
\usepackage{multirow}
\usepackage[super]{nth}
\usepackage[toc,page]{appendix}

\title{SALD: Sign Agnostic Learning with \\Derivatives} 

\author{Matan Atzmon \& Yaron Lipman  \\
Weizmann Institute of Science \\
\texttt{\{matan.atzmon,yaron.lipman\}@weizmann.ac.il} \\

}

\iclrfinalcopy % Uncomment for camera-ready version, but NOT for submission.
\begin{document}

\maketitle

\begin{abstract}
Learning 3D geometry directly from raw data, such as point clouds, triangle soups, or unoriented meshes is still a challenging task that feeds many downstream computer vision and graphics applications. 

In this paper, we introduce SALD: a method for learning implicit neural representations of shapes directly from raw data. We generalize sign agnostic learning (SAL) to include derivatives: given an unsigned distance function to the input raw data, we advocate a novel sign agnostic regression loss, incorporating both pointwise values and gradients of the unsigned distance function. Optimizing this loss leads to a \emph{signed} implicit function solution, the zero level set of which is a high quality and valid manifold approximation to the input 3D data. The motivation behind SALD is that incorporating derivatives in a regression loss leads to a lower sample complexity, and consequently better fitting. In addition, we prove that SAL enjoys a minimal length property in 2D, favoring minimal length solutions. More importantly, we are able to show that this property still holds for SALD, \ie,  with derivatives included.

We demonstrate the efficacy of SALD for shape space learning on two challenging datasets: ShapeNet \citep{chang2015shapenet} that contains inconsistent orientation and non-manifold meshes, and D-Faust \citep{bogo2017dynamic} that contains raw 3D scans (triangle soups). On both these datasets, we present state-of-the-art results.
\end{abstract}

\section{Introduction}
Recently, neural networks (NN) have been used for representing and reconstructing 3D surfaces. 
Current NN-based 3D learning approaches differ in two aspects: the choice of surface representation, and the supervision method. Common representations of surfaces include using NN as parameteric charts of surfaces \citep{groueix2018papier,williams2019deep}; volumetric implicit function representation defined over regular grids \citep{wu2016learning,tatarchenko2017octree,jiang2020sdfdiff}; and NN used directly as volumetric implicit functions \citep{Park_2019_CVPR,mescheder2019occupancy,atzmon2019controlling,chen2019learning}, referred henceforth as \emph{implicit neural representations}. 
Supervision methods include regression of known or approximated volumetric implicit representations \citep{Park_2019_CVPR,mescheder2019occupancy,chen2019learning}, regression directly with raw 3D data \citep{atzmon2019sal,gropp2020implicit,atzmon2019sal}, and differentiable rendering using 2D data (\ie, images) supervision \citep{niemeyer2019differentiable,liu2019learning,saito2019pifu,yariv2020multiview}. 

\begin{figure}[t]
    \centering
    \includegraphics[width=1.0\columnwidth]{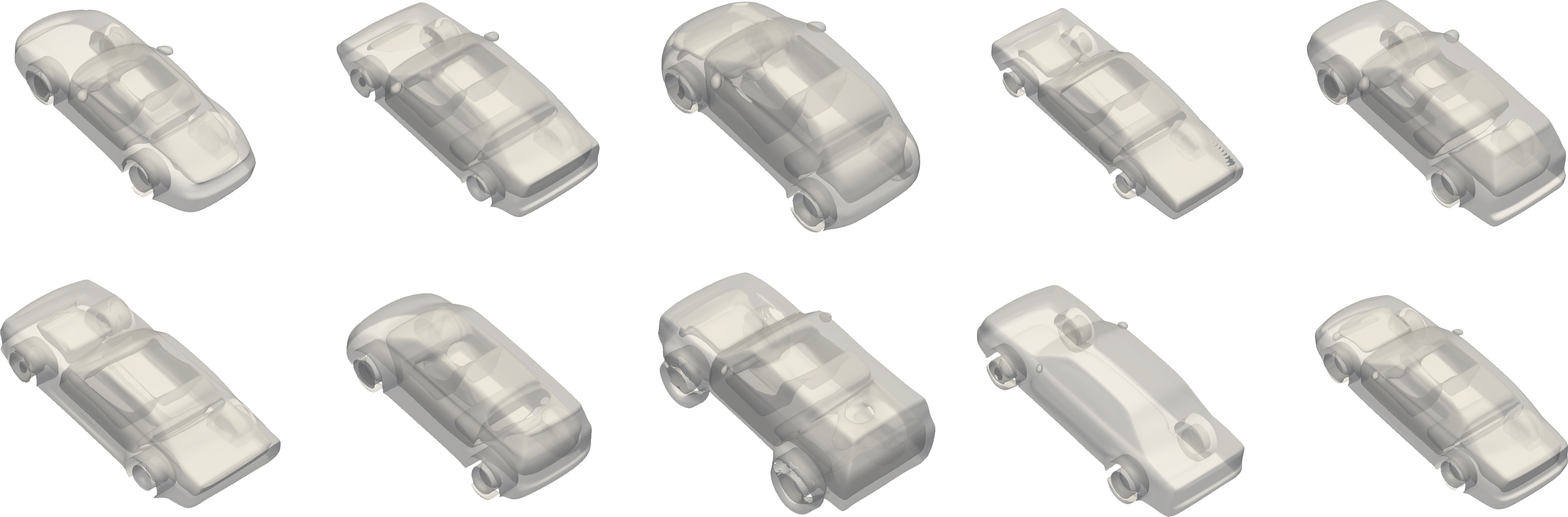} 
    \caption{Learning the shape space of ShapeNet \citep{chang2015shapenet} cars directly from raw data using \sald. Note the interior details; top row depicts \sald~reconstructions of train data, and bottom row \sald~reconstructions of test data.}
    \label{fig:teaser_cars_vae_train_test}
\end{figure}

The goal of this paper is to introduce \sald, a method for learning implicit neural representations of surfaces directly from \emph{raw 3D data}. 
The benefit in learning directly from raw data, \eg, non-oriented point clouds or triangle soups (\eg, \cite{chang2015shapenet}) and raw scans (\eg, \cite{bogo2017dynamic}), is avoiding the need for a ground truth signed distance representation of all train surfaces for supervision. This allows working with complex models with inconsistent normals and/or missing parts. In Figure \ref{fig:teaser_cars_vae_train_test} we show reconstructions of zero level sets of \sald~learned implicit neural representations of car models from the ShapeNet dataset \citep{chang2015shapenet} with variational auto-encoder; notice the high detail level and the interior, which would not have been possible with, \eg, previous data pre-processing techniques using renderings of visible parts \citep{Park_2019_CVPR}. 

Our approach improves upon the recent Sign Agnostic Learning (SAL) method \citep{atzmon2019sal} and shows that incorporating \emph{derivatives} in a sign agnostic manner provides a significant improvement in surface approximation and detail. 
SAL is based on the observation that given an unsigned distance function $h$ to some raw 3D data $\gX\subset\Real^3$, a sign agnostic regression to $h$ will introduce new local minima that are \emph{signed} versions of $h$; in turn, these signed distance functions can be used as implicit representations of the underlying surface. In this paper we show how the sign agnostic regression loss can be extended to compare both function values $h$ and \emph{derivatives} $\nabla h$, up to a sign. 

The main motivation for performing NN regression with derivatives is that it reduces the \emph{sample complexity} of the problem \citep{czarnecki2017sobolev}, leading to better accuracy and generalization. For example, consider a one hidden layer NN of the form $f(x) = \max\left\{ax,bx\right\} + c$. Prescribing two function samples at $\left\{-1,1\right\}$ are not sufficient for uniquely determining $f$, while adding derivative information at these points determines $f$ uniquely. 
%In practice, \sald~provides high level of detail approximation and, in contrast to previous work \ma{not true for SAL}, allows working directly with un-oriented data and normals, \eg, as is often the case with man made objects, and merged and/or noisy scans.

% Although the addition of derivative loss does not provide extra (\ie, previously unknown) information on the function $h$ it nevertheless leads to significantly better signed distance functions that capture more detail than the original SAL. 

Analyzing theoretical aspects of \sal~and \sald, we observe that both possess the favorable minimal surface property, that is, in areas of missing parts and holes they will prefer zero level sets with minimal area. We justify this property by proving that, in 2D, when restricted to the zero level-set (a curve in this case), the \sal~and \sald~losses  would encourage a straight line solution connecting neighboring data points. 

We have tested \sald~on the dataset of man-made models, ShapeNet \citep{chang2015shapenet}, and human raw scan dataset, D-Faust \citep{bogo2017dynamic}, and compared to state-of-the-art methods. In all cases we have used the raw input data $\gX$ as is and considered the unsigned distance function to $\gX$, \ie, $h_\gX$, in the \sald~loss to produce an approximate signed distance function in the form of a neural network. Comparing to state-of-the-art methods we find that SALD achieves superior results on this dataset.
On the D-Faust dataset, when comparing to ground truth reconstructions we report state-of-the-art results, striking a balance between approximating details of the scans and avoiding overfitting noise and ghost geometry. \\

\noindent Summarizing the contributions of this paper:
\begin{itemize}[label=$\bullet$]
    \item Introducing sign agnostic learning with derivatives. 
    \item Identifying and providing a theoretical justification for the minimal surface property of sign agnostic learning. 
    \item Training directly on raw data (end-to-end) including unoriented or not consistently oriented triangle soups and raw 3D scans. 
\end{itemize}

\section{Previous work}
Learning 3D shapes with neural networks and 3D supervision has shown great progress recently. We review related works, where we categorize the existing methods based on their choice of 3D surface representation. 

\textbf{Parametric representations.}
The most fundamental surface representation is an \emph{atlas}, that is a collection of parametric charts $f:\Real^2\too\Real^3$ with certain coverage and transition properties \citep{do2016differential}. \cite{groueix2018papier} adapted this idea using neural network to represent a surface as union of such charts; \cite{williams2019deep} improved this construction by introducing better transitions between charts;  
\cite{sinha2016deep} use geometry images \citep{gu2002geometry} to represent an entire shape using a single chart; \cite{maron2017convolutional} use global conformal parameterization for learning surface data; \cite{ben2018multi} use a collection of overlapping global conformal charts for human-shape generative model.
The benefit in parametric representations is in the ease of sampling the learned surface (\ie, forward pass) and work directly with raw data (\eg, Chamfer loss); their main struggle is in producing charts that are collectively consistent, of low distortion, and covering the shape. 

\textbf{Implicit representations.}
Another approach for representing surfaces is as zero level sets of a function, called an \emph{implicit function}. There are two popular methods to model implicit volumetric functions with neural networks: i) \emph{Convolutional neural network} predicting scalar values over a predefined fixed volumetric structure (\eg, grid or octree) in space \citep{tatarchenko2017octree,wu2016learning}; and ii) \emph{Multilayer Perceptron} of the form $f:\Real^3\too\Real$ defining a continuous volumetric function \citep{Park_2019_CVPR,mescheder2019occupancy,chen2019learning}. 
Currently, neural networks are trained to be implicit function representations with two types of supervision: (i) regression of samples taken from a known or pre-computed implicit function representation such as occupancy function \citep{mescheder2019occupancy,chen2019learning} or a signed distance function \citep{Park_2019_CVPR}; and (ii) working with raw 3D supervision, by particle methods relating points on the level sets to the model parameters \citep{atzmon2019controlling}, using sign agnostic losses \citep{atzmon2019sal}, or supervision with PDEs defining signed distance functions \citep{gropp2020implicit}.

\textbf{Primitives.}
Another type of representation is to learn shapes as composition or unions of a family of \emph{primitives}. 
\cite{li2019supervised} represent a shape using a parametric collection of primitives. 
\cite{genova2019learning,genova2019deep} use a collection of Gaussians and learn consistent shape decompositions. 
\cite{chen2019bsp} suggest a differentiable Binary Space Partitioning tree (BSP-tree) for representing shapes. 
\cite{deprelle2019learning} combine points and charts representations to learn basic shape structures. \cite{deng2019cvxnets} represent a shape as a union of convex sets. \cite{williams2020voronoinet} learn cites of Voronoi cells for implicit shape representation.  

\textbf{Template fitting.}
Lastly, several methods learn 3D shapes of a certain class (\eg, humans) by learning the deformation from a template model. Classical methods use matching techniques and geometric loss minimization for non-rigid template matching \citep{allen2002articulated,allen2003space,anguelov2005scape}. \cite{groueix20183d} use an auto-encoder architecture and Chamfer distance to match target shapes. \cite{litany2018deformable} use graph convolutional autoencoder to learn deformable template for shape completion.

\section{Method}
Given raw geometric input data $\gX\subset\Real^3$, \eg, a triangle soup, our goal is to find a multilayer perceptron (MLP) $f:\Real^3\times\Real^m\too\Real$ whose zero level-set, 
\begin{equation}\label{e:zero_level_set}
    \gS = \set{\vx\in\Real^3 \ \vert \ f(\vx;\theta)=0}
\end{equation}
is a manifold surface that approximates $\gX$. 

\textbf{Sign agnostic learning.} Similarly to SAL, our approach is to consider the (readily available) \emph{unsigned} distance function to the raw input geometry, 
\begin{equation}\label{e:h}
 h(\vy)=\min_{\vx\in\gX}\norm{\vy-\vx}   
\end{equation}
and perform sign agnostic regression to get a \emph{signed} version $f$ of $h$. SAL uses a loss of the form 
\begin{equation}\label{e:loss_sal}
    \loss(\theta) = \E_{\vx\sim \gD}\ \tau \big( f(\vx;\theta) , h(\vx) \big),
\end{equation}
where $\gD$ is some probability distribution, and $\tau$ is an unsigned similarity. That is, $\tau(a,b)$ is measuring the difference between scalars $a,b\in\Real$ up-to a sign. For example 
\begin{equation}\label{e:tau_sal}
 \tau(a,b)=\big||a|-b   \big| 
\end{equation}
is an example that is used in \cite{atzmon2019sal}. The key property of the sign agnostic loss in \eqref{e:loss_sal} is that, with proper weights initialization $\theta_0$, it finds a new \emph{signed} local minimum $f$ which in absolute value is similar to $h$. In turn, the zero level set $\gS$ of $f$ is a valid manifold describing the data $\gX$.

\textbf{Sign agnostic learning with derivatives.} Our goal is to generalize the SAL loss (\eqref{e:loss_sal}) to include derivative data of $h$ and show that optimizing this loss provides implicit neural representations, $\gS$, that enjoy better approximation properties with respect to the underlying geometry $\gX$. 

Generalizing \eqref{e:loss_sal} requires designing an unsigned similarity measure $\tau$ for vector valued functions. The key observation is that \eqref{e:tau_sal} can be written as  $\tau(a,b)=\min\set{\abs{a-b},\abs{a+b}}$, $a,b\in\Real$, and can be generalized to vectors $\va,\vb\in\Real^d$ by  
\begin{equation}\label{e:tau_norm_diff}
    \tau(\va,\vb)=\min\set{ \norm{\va-\vb} , \norm{\va+\vb} }.
\end{equation}
% where $\norm{\cdot}$ is some norm; we use the $L_2$ norm, $\norm{\va}=\norm{\va}_2=\sqrt{\va^T\va}$. A second option is 
% \begin{equation}\label{e:tau_sin}
%     \tau(\va,\vb)=\abs{\sin{\alpha}}
% \end{equation}
% where $\alpha=\sphericalangle(\va,\vb)$ the angle between $\va,\vb$. Note that $\tau(-\va,\vb)=\abs{\sin{(\pi-\alpha)}}=\abs{\sin{\alpha}}=\tau(\va,\vb)$ so $\tau$ is unsigned. The difference between the two options of $\tau$ is that \eqref{e:tau_sin} only penalizes the difference in directions of $\va,\vb$, while \eqref{e:tau_norm_diff} penalizes both angle and length differences. 

We define the \sald~loss:
\begin{equation}\label{e:loss_sald}
\loss(\theta) = \E_{\vx\sim \gD}\ \tau \big( f(\vx;\theta) , h(\vx) \big) + \lambda \E_{\vx\sim \gD'}\ \tau \big( \nabla_\vx f(\vx;\theta) , \nabla_\vx h(\vx) \big)
\end{equation}
where $\lambda>0$ is a parameter, $\gD'$ is a probability distribution, and $\nabla_\vx f(\vx;\theta),\nabla_\vx h(\vx)$ are the gradients $f,h$ (resp.) with respect to their input $\vx$. 

\begin{figure}[t]
    \centering
    \begin{tabular}{ccc}
         \includegraphics[width=0.3\columnwidth]{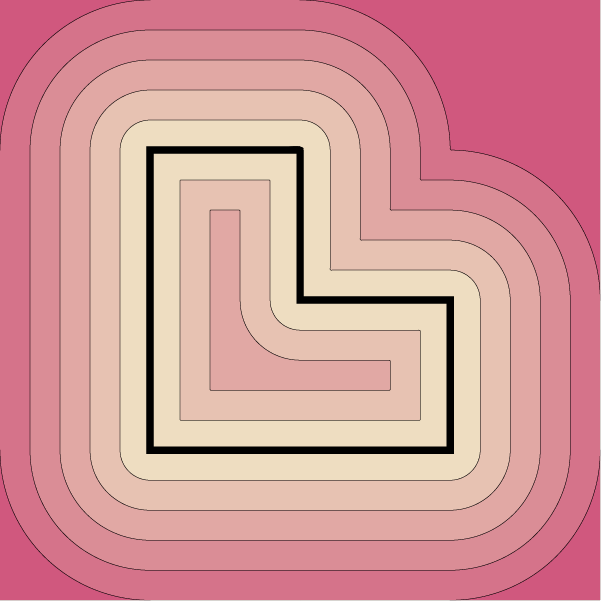} &  
         \includegraphics[width=0.3\columnwidth]{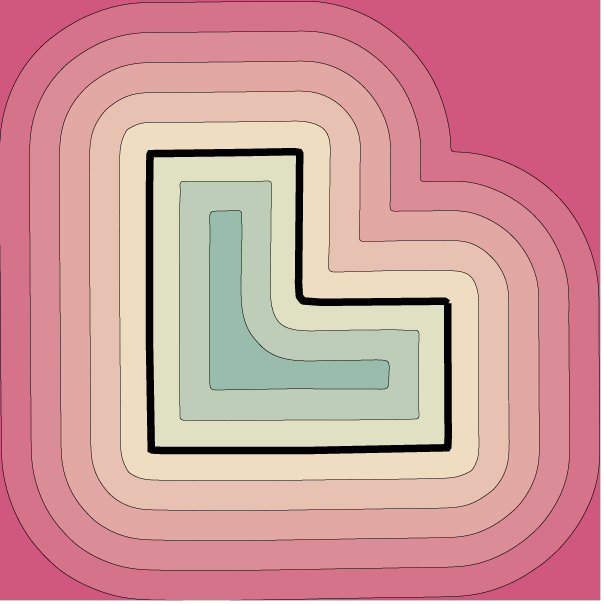} &
         \includegraphics[width=0.3\columnwidth]{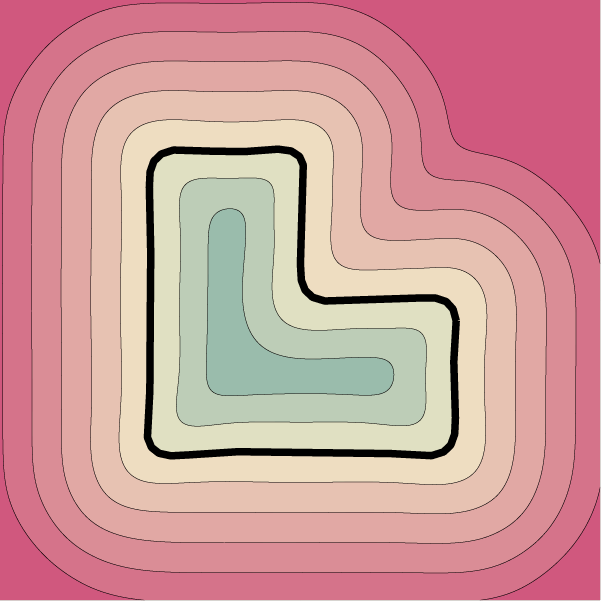} \\
         unsigned distance & \sald & \sal
    \end{tabular}
    \caption{Sign agnostic learning of an unsigned distance function to an L shape (left). Red colors depict positive values, and blue-green colors depict negative values. In the middle, the result of optimizing the \sald~loss (\eqref{e:loss_sald}); on the right, the result of \sal~loss (\eqref{e:loss_sal}). Note that \sald~better preserves sharp features of the shape and the isolevels.}
    \label{fig:L}
\end{figure}

In Figure \ref{fig:L} we show the unsigned distance $h$ to an L-shaped curve (left), and the level sets of the MLPs optimized with the \sald~loss (middle) and the \sal~loss (right); note that \sald~loss reconstructed the sharp features (\ie, corners) of the shape and the level sets of $h$, while \sal~loss smoothed them out; the implementation details of this experiment can be found in Appendix \ref{appendix:figs}.  %\yl{how we chose $D$? how many epochs? 5$k$}  

\textbf{Minimal surface property.}\label{p:minimal} We show that the \sal~and \sald~losses possess a \emph{minimal surface property} \citep{zhao2001fast}, that is, they strives to minimize surface area of missing parts. For example, Figure \ref{fig:U} shows the unsigned distance to a curve with a missing segment (left), and the zero level sets of MLPs optimized with \sald~loss (middle), and \sal~loss (right). Note that in both cases the zero level set in the missing part area is the minimal length curve (\ie, a line) connecting the end points of that missing part. \sald~also preserves sharp features of the rest of the shape.

We will provide a theoretical justification to this property in the 2D case. We consider a geometry defined by two points in the plane, $\gX=\set{\vx_1,\vx_2}\subset \Real^2$ and possible solutions where the zero level set curve $\gS$ is connecting $\vx_1$ and $\vx_2$. 
We prove that among a class of curves $\gU$ connecting $\vx_1$ and $\vx_2$, the straight line minimizes the losses in \eqref{e:loss_sal} and \eqref{e:loss_sald} restricted to $\gU$, when assuming uniform distributions $\gD,\gD'$.
\begin{wrapfigure}[9]{r}{0.3\textwidth}
  \begin{center}\vspace{-5pt}
    \includegraphics[width=0.28\textwidth]{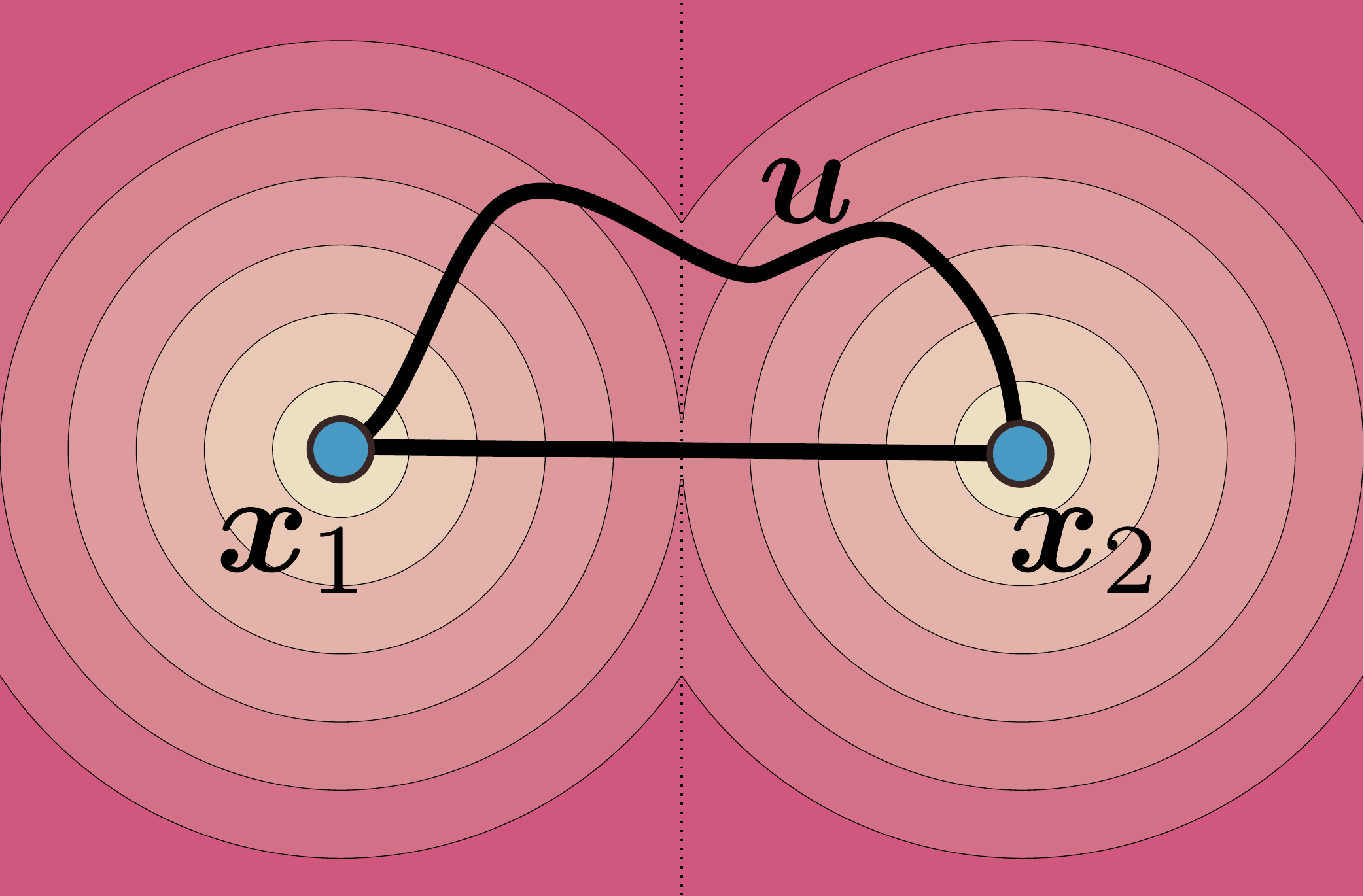}
  \end{center}\vspace{-16pt}
  \caption{Minimal surface property in 2D.}\label{fig:minimal_curve}
\end{wrapfigure}
We assume (without losing generality) that $\vx_1=(0,0)^T$, $\vx_2=(\ell,0)^T$ and consider curves $\vu\in\gU$ defined by $\vu(s)=(s,t(s))^T$, where $s\in [0,\ell]$, 
and $t:\Real\too\Real$ is some differentiable function such that $t(0)=0=t(\ell)$, see Figure \ref{fig:minimal_curve}.

For the \sald~loss we prove the claim for a slightly simplified agnostic loss motivated by the following lemma proved in Appendix \ref{appendix:proof}: 
\begin{replemma}{lem:sin}
For any pair of unit vectors $\va,\vb$: $\min\set{\norm{\va-\vb}, \norm{\va+\vb}}\geq \abs{\sin \angle(\va,\vb)}$. 
\end{replemma}
We consider $\tau(\va,\vb)=\abs{\sin \angle(\va,\vb)}$ for the derivative part of the loss in \eqref{e:loss_sald}, which is also sign agnostic. 
\begin{theorem}
Let $\gX=\set{\vx_1,\vx_2}\subset \Real^2$, and the family of curves $\gU$ connecting $\vx_1$ and $\vx_2$. Furthermore, let $\loss_{\scriptscriptstyle\text{\sal}}(\vu)$ and $\loss_{\scriptscriptstyle\text{\sald}}(\vu)$ denote the losses in \eqref{e:loss_sal} and \eqref{e:loss_sald} (resp.) when restricted to $\vu$ with uniform distributions $\gD,\gD'$. Then in both cases the straight line, \ie, the curve $\vu(s)=(s,0)$, is the strict global minimizer of these losses.
\end{theorem}
\begin{proof}
The unsigned distance function is $$h(\vu)=\begin{cases}\sqrt{s^2+t^2} & s\in [0,\ell/2] \\ \sqrt{(s-\ell)^2+t^2} & s\in (\ell/2,\ell] \end{cases}.$$
From symmetry it is enough to consider only the first half of the curve, \ie, $s\in[0,\ell/2)$.
Then, the \sal~loss, \eqref{e:loss_sal}, restricted to the curve $\vu$ (\ie, where $f$ vanishes) takes the form $$\loss_{\scriptscriptstyle\text{\sal}}(\vu)=\int_0^{\ell/2} \tau(f(\vu;\theta),h(\vu))\norm{\dot{\vu}}\ ds=\int_0^{\ell/2} \sqrt{s^2+t^2}\sqrt{1+\dot{t}^2}\ ds,$$ where $\sqrt{1+\dot{t}^2} \ ds$ is the length element on the curve $\vu$, and $\tau(f(s,t;\theta),h(s,t))=\abs{h(s,t)}=\sqrt{s^2+t^2}$, since $f(s,t;\theta)=0$ over the curve $\vu$. Plugging $t(s)\equiv 0$ in $\loss_{\scriptscriptstyle \text{\sal}}(\vu)$ we see that the curve $\vu=(s,0)^T$, namely the straight line curve from $\vx_1$ to $0.5(\vx_1+\vx_2)$ is a strict global minimizer of $\loss_{\scriptscriptstyle\text{\sal}}(\vu)$. Similar argument on $s\in[\ell/2,\ell]$ prove the claim for the \sal~case. 

\begin{figure}[t]
    \centering
    \begin{tabular}{ccc}
          \includegraphics[width=0.3\columnwidth]{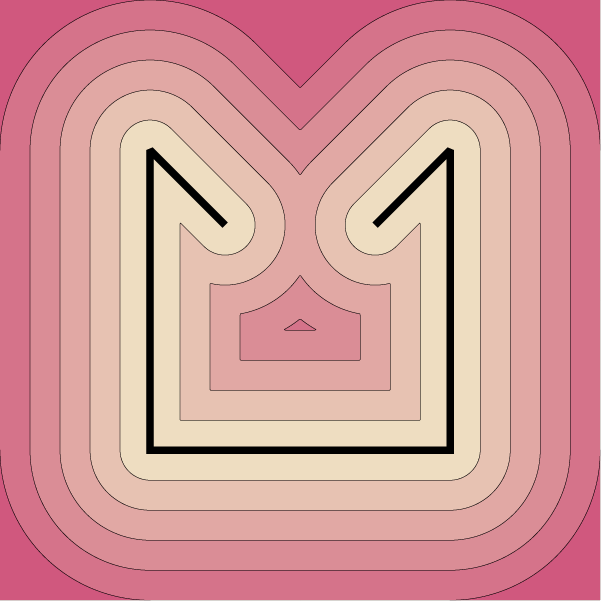} &
          \includegraphics[width=0.3\columnwidth]{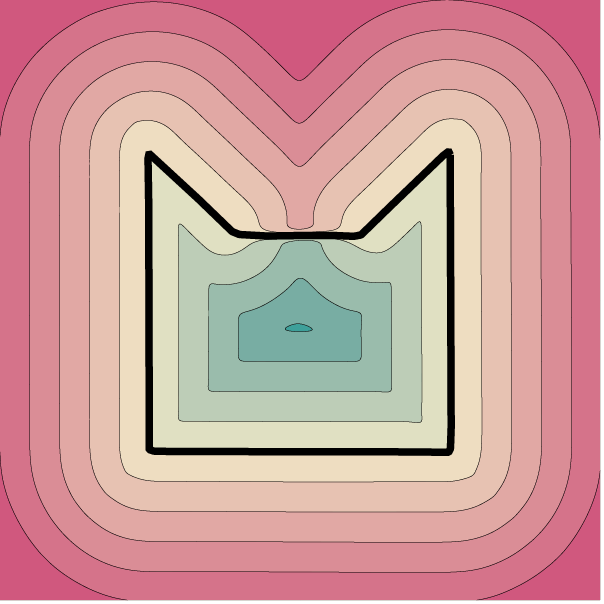} &
          \includegraphics[width=0.3\columnwidth]{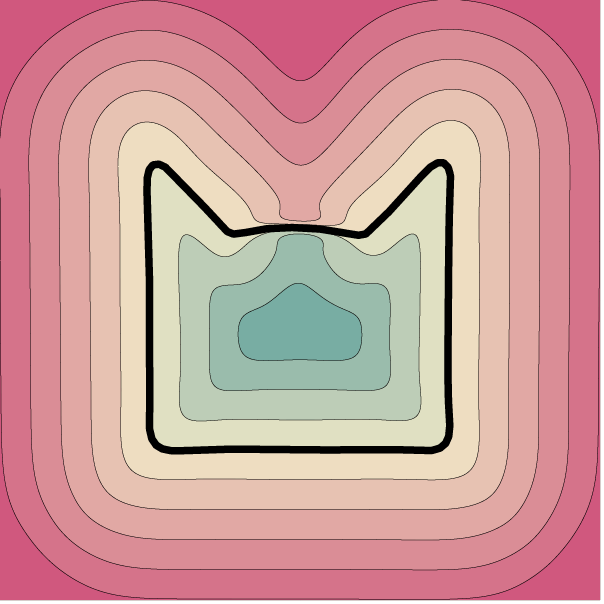} \\
         unsigned distance & \sald & \sal
    \end{tabular}
    \caption{Minimal surface property: using \sald~(middle) and \sal~(right) with the input unsigned distance function of a curve with a missing part (left) leads to a solution (black line, middle and right) with approximately minimal length in the missing part area. Note that the \sald~solution also preserves sharp features of the original shape, better than \sal. }
    \label{fig:U}
\end{figure}

For the \sald~case, we want to calculate $\tau(\nabla_\vx f(\vu;\theta), \nabla_\vx h(\vu))$ restricted to the curve $\vu$; let $\va=\nabla_\vx f(\vu;\theta)$ and $\vb = \nabla_\vx h(\vu)$. First, $\vb = (s^2+t^2)^{-1/2}(s,t)^T$. Second, $\va$ is normal to the curve $\vu$, therefore it is proportional to $\dot{\vu}^\perp=(-\dot{t},1)^T$. Next, note that
$$\abs{\sin \angle(\va,\vb)} = \frac{\abs{\det \begin{pmatrix}-\dot{t}&s\\ 1& t \end{pmatrix}}}{\sqrt{1+\dot{t}^2}\sqrt{s^2+t^2}}=\frac{1}{\sqrt{1+\dot{t}^2}}\abs{\frac{d}{ds}\norm{(s,t)}},$$
where the last equality can be checked by differentiating $\norm{(s,t)}$ w.r.t.~$s$. Therefore,
\begin{align*}
\frac{
  \loss_{\scriptscriptstyle\text{\sald}}(\vu)-\loss_{\scriptscriptstyle\text{\sal}}(\vu)}{\lambda}&=\int_0^{\ell/2} \tau(\va,\vb)\norm{\dot{\vu}} ds = \int_0^{\ell/2}  \abs{\frac{d}{ds}\norm{(s,t)}} ds \geq  \norm{\parr{\frac{\ell}{2},t\parr{\frac{\ell}{2}}}} \geq \frac{ \ell}{2}.
\end{align*}
This bound is achieved for the curve $\vu=(s,0)$, which is also a minimizer of the \sal~loss. The straight line also minimizes this version of the \sald~loss since $\loss_{\scriptscriptstyle\text{\sald}}(\vu) = \parr{ \loss_{\scriptscriptstyle\text{\sald}}(\vu)-\loss_{\scriptscriptstyle\text{\sal}}(\vu)} + \loss_{\scriptscriptstyle\text{\sal}}(\vu).$ %\qed
\end{proof}

\section{Experiments}\label{s:exps}
We tested \sald~on the task of shape space learning from raw 3D data. We experimented with two different datasets: i) ShapeNet dataset \citep{chang2015shapenet}, containing synthetic 3D Meshes; and ii) D-Faust dataset \citep{bogo2017dynamic} containing raw 3D scans. 

\textbf{Shape space learning architecture.}
%\textbf{Architecture.}
Our method can be easily incorporated into existing shape space learning architectures: i) Auto-Decoder (AD) suggested in \cite{Park_2019_CVPR}; and the ii) Modified Variational Auto-Encoder (VAE) used in \cite{atzmon2019sal}. For VAE, the encoder is taken to be PointNet \citep{qi2017pointnet}. For both options, the decoder is the implicit representation in \eqref{e:zero_level_set}, where $f(\vx;\theta)$ is taken to be an 8-layer MLP with 512 hidden units in each layer and Softplus activation. In addition, to enable sign agnostic learning we initialize the decoder weights, $\theta$, using the geometric initialization from \cite{atzmon2019sal}. See Appendix \ref{appendix:arch} for more details regarding the architecture.

\textbf{Baselines.} The baseline methods selected for comparison cover both existing supervision methodologies: DeepSDF \citep{Park_2019_CVPR} is chosen as a representative out of the methods that require pre-computed implicit representation for training. For methods that train directly on raw 3D data, we compare versus SAL \citep{atzmon2019sal} and IGR \citep{gropp2020implicit}. See Appendix \ref{appendix:eval} for a detailed description of the quantitative metrics used for evaluation.

\begin{table}[]
\resizebox{\textwidth}{!}{%
\begin{tabular}{l|ll|ll|ll|ll|ll}
Category            & \multicolumn{2}{l}{Sofas}       & \multicolumn{2}{l}{Chairs}      & \multicolumn{2}{l}{Tables}      & \multicolumn{2}{l}{Planes}      & \multicolumn{2}{l}{Lamps}       \\ 
                    & Mean           & Median         & Mean           & Median         & Mean           & Median         & Mean           & Median         & Mean           & Median         \\ \hline
DeepSDF              & \color{blue}{\textbf{0.329}}          & \color{blue}{\textbf{0.230}}          & \color{blue}{\textbf{0.341}}          & \textbf{0.133} & 0.839          & \textbf{0.149} & \color{blue}{\textbf{0.177}}          & 0.076          & \color{blue}{\textbf{0.909}}          & \color{blue}{\textbf{0.344}}          \\
\sal     & 0.704          & 0.523         & 0.494          & 0.259          & \color{blue}{\textbf{0.543}}          & \color{blue}{\textbf{0.231}}          & 0.429          & 0.146          & 4.913          & 1.515          \\
\sald (VAE)         & 0.391          & 0.244          & 0.415          & 0.255          & 0.679          & 0.279          & 0.197          & \color{blue}{\textbf{0.062}}          & 1.808          & 1.172          \\
\sald (AD) & \textbf{0.207} & \textbf{0.147} & \textbf{0.281} & \color{blue}{\textbf{0.157}}          & \textbf{0.408} & 0.25           & \textbf{0.098} & \textbf{0.032} & \textbf{0.506} & \textbf{0.327}\vspace{5pt}
\end{tabular}}
\caption{ShapeNet quantitative results. We log the mean and median of the Chamfer distances ($\dist_{\text{C}}$) between the reconstructed 3D surfaces and the ground truth meshes. Numbers are reported $* 10^3$.} %\vspace{-15pt}
    \label{tab:shapenet_test}
\end{table}
\begin{figure}[t]
    \centering
    \includegraphics[width=1.0\columnwidth]{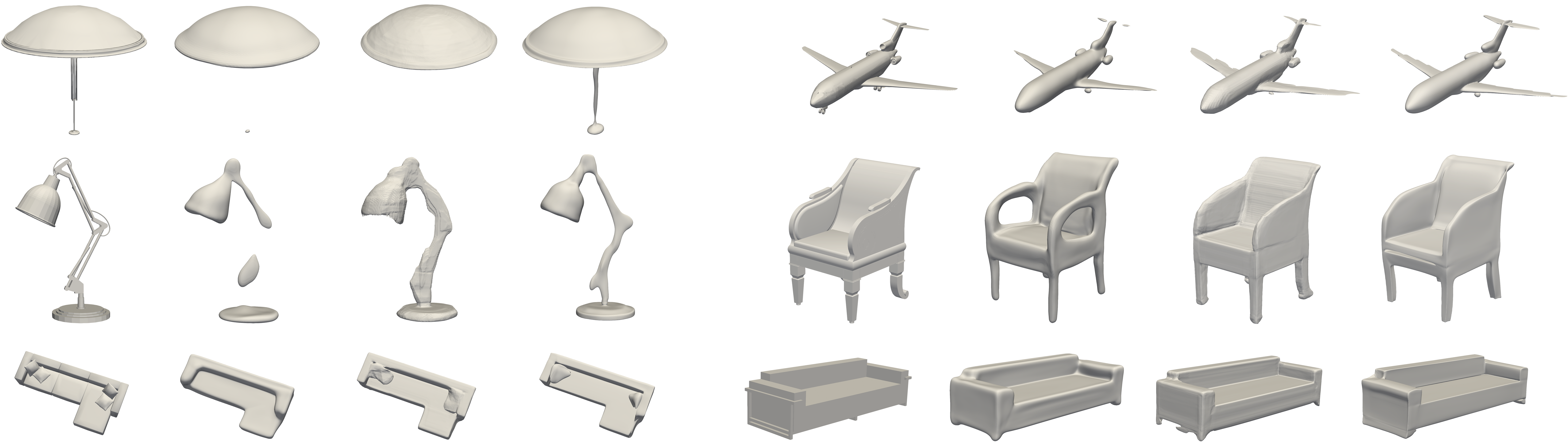}
    \caption{ShapeNet qualitative test results. Each quadruple shows (columns from left to right): ground truth model, \sal-reconstruction, DeepSDF reconstruction, \sald~reconstruction. }
    \label{fig:shapenet}
\end{figure}
\subsection{ShapeNet}
In this experiment we tested \sald~ability to learn a shape space by training on a challenging 3D data such as non-manifold/non-orientable meshes. We tested \sald~with both AD and VAE architectures. In both settings, we set $\lambda=0.1$ for the \sald~loss.
We follow the evaluation protocol as in \cite{Park_2019_CVPR}: using the same train/test splits, we train and evaluate our method on $5$ different categories. Note that comparison versus IGR is omitted as IGR requires consistently oriented normals for shape space learning, which is not available for ShapeNet, where many models have non-consistent triangles' orientation.

\textbf{Results.}
Table \ref{tab:shapenet_test} and Figure \ref{fig:shapenet} show quantitative and qualitative results (resp.) for the held-out test set, comparing \sal, DeepSDF and \sald.
As can be read from the table and inspected in the figure, our method, when used with the same auto-decoder as in DeepSDF, compares favorably to DeepSDF's reconstruction performance on this data.

\begin{wrapfigure}[7]{r}{0.52\textwidth}
  \begin{center}\vspace{-14pt}
    \includegraphics[width=0.5\textwidth]{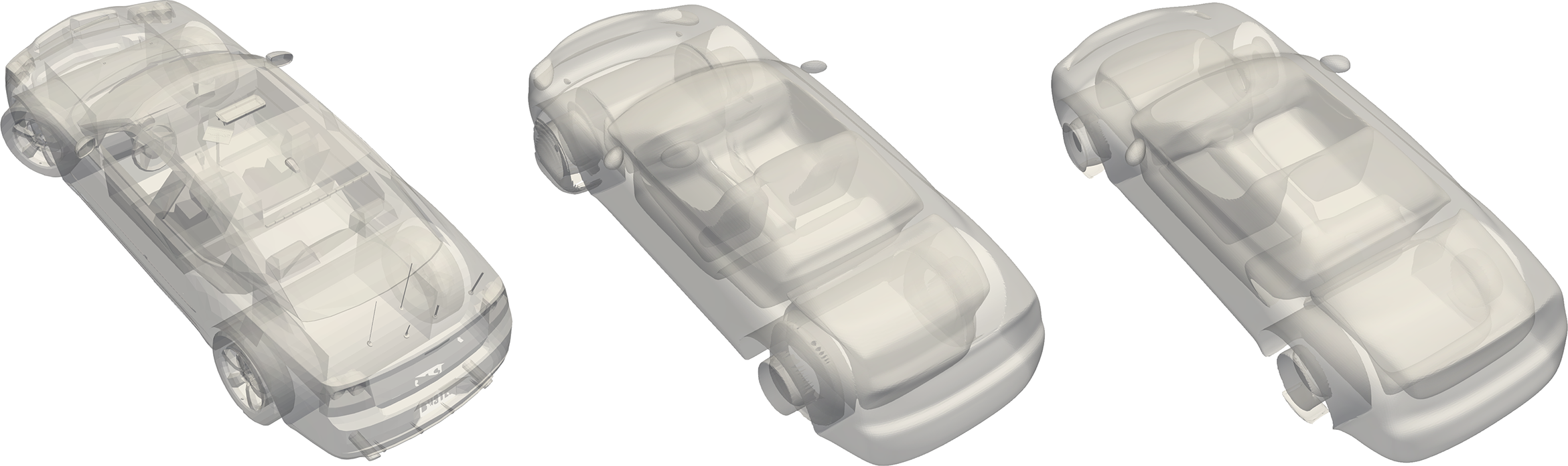}\vspace{-10pt}
  \end{center}
  \vspace{-5pt}
  \caption{AD versus VAE.}\label{fig:ad_vs_vae}
\end{wrapfigure}
\begin{figure}[h]
    \centering
    \includegraphics[width=1.0\columnwidth]{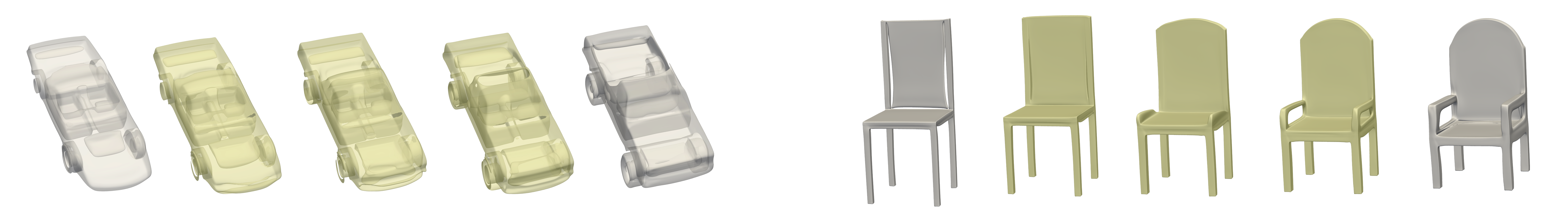} 
    \caption{ShapeNet latent interpolation. In each group, the leftmost and rightmost columns are test examples reconstructions; latent space generated shapes are coloured in yellow.}
    \label{fig:shapenet_latent}
\end{figure}

Qualitatively the surfaces produces by \sald~are smoother, mostly with more accurate sharp features, than \sal~and DeepSDF generated surfaces. Figure \ref{fig:teaser_cars_vae_train_test} shows typical train and test results from the Cars class with VAE. Figure \ref{fig:ad_vs_vae} shows a comparison between \sald~shape space learning with VAE and AD in reconstruction of a test car model (left). Note that the AD (middle) seems to produce more details of the test model than the VAE (right), \eg, steering wheel and headlights. Figure \ref{fig:shapenet_latent} show \sald~(AD) generated shapes via latent space interpolation between two test models.

% \begin{figure}[t]
%     \centering
%     \includegraphics[width=1.0\columnwidth]{figures/ad_versus_vae.png} 
%     \caption{Comparing reconstruction of a test car model (left) with AD \sald~(middle), and VAE \sald~(right). Note the auto-decoder is able to produce more details of the test model, \eg steering wheel and headlights.  }
%     \label{fig:ad_vs_vae}
% \end{figure}

\subsection{D-Faust}
\begin{figure}[t]
    \centering
    \includegraphics[width=1.0\columnwidth]{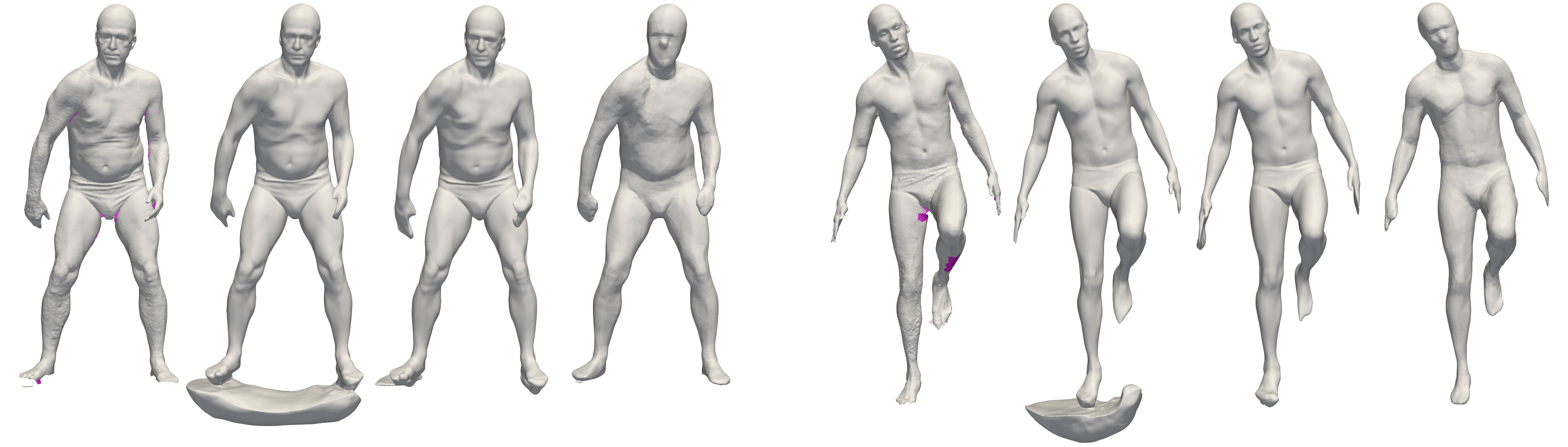} 
    \caption{D-Faust qualitative results on test examples.  Each quadruple shows (columns from left to right): raw scans (magenta depict back-faces), IGR, \sald, and \sal. }
    \label{fig:dfaust}
\end{figure}

The D-Faust dataset \citep{bogo2017dynamic} contains raw scans (triangle soups) of 10 humans in multiple poses. There are approximately 41k scans in the dataset. Due to the low variety between adjacent scans, we sample each pose scans at a ratio of $1:5$. The leftmost column in Figure \ref{fig:dfaust} shows examples of raw scans used for training. For evaluation we use the \emph{registrations} provided with the data set. Note that the registrations where not used for training. We tested \sald~using the VAE architecture, with $\lambda=1.0$ set for the \sald~loss. We followed the evaluation protocol as in \cite{atzmon2019sal}, using  the same train/test split. Note that \cite{atzmon2019sal} already conducted a comprehensive comparison of \sal~versus DeepSDF, establishing \sal~as a state-of-the-art method for this dataset. Thus, we focus on comparison versus \sal~and IGR.

\begin{figure}[h]
    \centering
    \includegraphics[width=1.0\columnwidth]{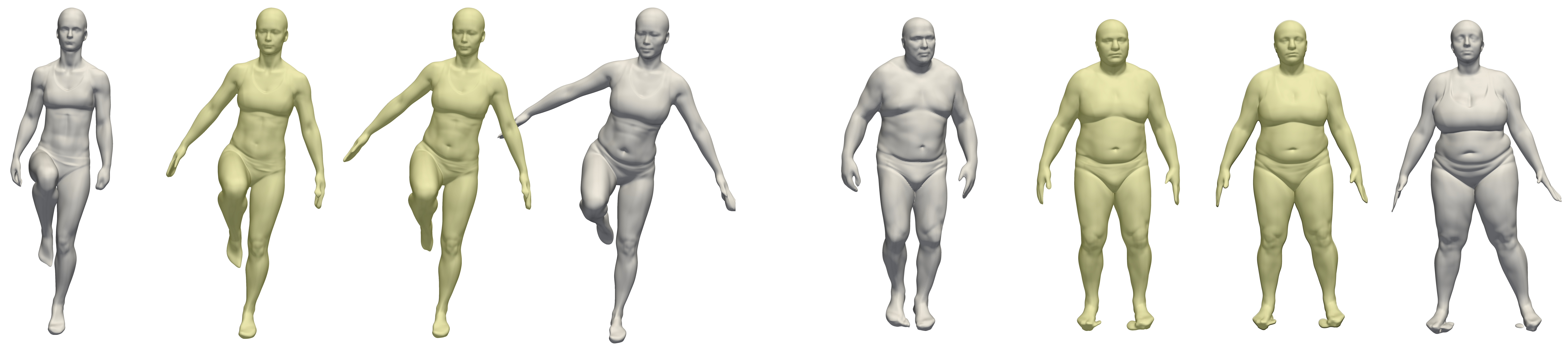} 
    \caption{D-Faust latent interpolation. In each group, the leftmost and rightmost columns are test scans reconstructions; latent space generated shapes are coloured in yellow.}
    \label{fig:dfaust_latent}
\end{figure}

\textbf{Results.}
Table \ref{tab:dfaust_train} and Figure \ref{fig:dfaust} show quantitative and qualitative results (resp.); although \sald~does not produces the best test quantitative results, it is roughly comparable in every measure to the best among the two baselines. That is, it produces details comparable to IGR while maintaining the minimal surface property as \sal~and not adding undesired surface sheets as IGR; see the figure for visual illustrations of these properties: the high level of details of \sald~and IGR compared to \sal, and the extraneous parts added by IGR, avoided by \sald. These phenomena can also be seen quantitatively, \eg, the reconstruction-to-registration loss of IGR. Figure \ref{fig:dfaust_latent} show \sald~generated shapes via latent space interpolation between two test scans. Notice the ability of \sald~to generate novel mixed faces and body parts.

\begin{table}[b]
\resizebox{\textwidth}{!}{%
\begin{tabular}{l|llll|llll|llll}
                              & \multicolumn{2}{l}{$\dist_{\text{C}}^{\rightarrow}\left(\text{reg.},\text{recon.}\right) $  } & \multicolumn{2}{l}{$\dist_{\text{N}}^{\rightarrow}\left(\text{reg.},\text{recon.}\right) $  }                              & \multicolumn{2}{l}{$\dist_{\text{C}}^{\rightarrow}\left(\text{recon.},\text{reg.}\right) $}      &     \multicolumn{2}{l}{$\dist_{\text{N}}^{\rightarrow}\left(\text{recon.},\text{reg.}\right) $  }                       & \multicolumn{2}{l}{$\dist_{\text{C}}^{\rightarrow}\left(\text{scan},\text{recon.}\right) $} & \multicolumn{2}{l}{$\dist_{\text{N}}^{\rightarrow}\left(\text{scan},\text{recon.}\right) $}                               \\
                              & Mean           & Median         & Mean            & Median          & Mean           & Median         & Mean            & Median          & Mean           & Median         & Mean           & Median         \\ \hline
% \multirow{3}{*}{Train} & SAL   & 0.261          & \textbf{0.239} & 12.303          & 12.122          & \textbf{0.286} & \textbf{0.188} & 10.41           & 10.88           & 0.188          & 0.175          & 9.038          & 8.825          \\
%                       & IGR   & \textbf{0.256} & 0.242          & \textbf{10.271} & \textbf{10.166} & 3.628          & 3.502          & 16.894          & 17.507          & 0.258          & 0.169          & \textbf{5.802} & \textbf{5.622} \\
%                       & \sald & 0.262          & 0.242          & 10.455          & 10.303          & 0.714          & 0.357          & \textbf{10.178} & \textbf{10.475} & \textbf{0.184} & \textbf{0.173} & 6.077          & 5.94           \\
%                       \hline
   SAL   & \color{blue}{\textbf{0.418}}          & \color{blue}{\textbf{0.328}}          & 13.21          & 12.459          & \textbf{0.344} & \textbf{0.256} & \color{blue}{\textbf{11.354}} & \color{blue}{\textbf{10.522}} & 0.429          & \color{blue}{\textbf{0.246}}          &       10.096         &    9.096            \\
                        IGR   & \textbf{0.276} & \textbf{0.187} & \textbf{10.328}  & \textbf{9.822}  & 3.806           & 3.627          & 17.124           & 17.902          & \textbf{0.241}  & \textbf{0.11} &   \textbf{5.829}             &         \textbf{5.295}       \\
                        \sald & 0.428          & 0.346          & \color{blue}{\textbf{11.67}}          & \color{blue}{\textbf{11.07}}          & \color{blue}{\textbf{0.489}}          & \color{blue}{\textbf{0.362}}          & \textbf{11.035}          & \textbf{10.371}          & \color{blue}{\textbf{0.397}}          & 0.279          &  \color{blue}{\textbf{7.884}}              & \color{blue}{\textbf{7.227}}
\end{tabular}}\vspace{5pt}
\caption{D-Faust quantitative results. We log mean and median of the one-sided Chamfer and normal distances between registration meshes (reg), reconstructions (recon) and raw input scans (scan). The $\dist_{\text{C}}$ numbers are reported $*10^2$.} %\vspace{-15pt}
    \label{tab:dfaust_train}
\end{table}

\subsection{Limitations}
\begin{wrapfigure}[3]{r}{0.52\textwidth}
  \begin{center}\vspace{-45pt}
    \includegraphics[width=0.5\textwidth]{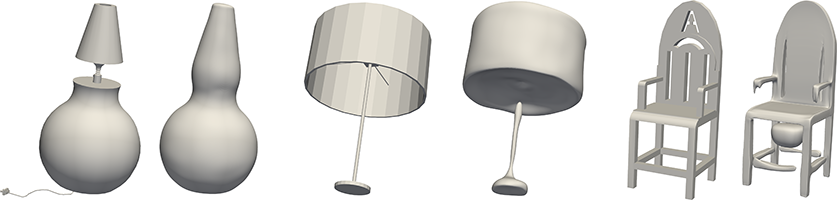}\vspace{-10pt}
  \end{center}
  \caption{Failure cases.}\label{fig:failures}
\end{wrapfigure}
 Figure \ref{fig:failures} shows typical failure cases of our method from the ShapeNet experiment described above.

We mainly suffer from two types of failures: First, since inside and outside information is not known (and often not even well defined in ShapeNet models) \sald~can add surface sheets closing what should be open areas (\eg, the bottom side of the lamp, or holes in the chair). Second, thin structures can be missed (\eg, the electric cord of the lamp on the left).

% \begin{figure}[t]
%     \centering
%     \includegraphics[width=1.0\columnwidth]{figures/failures.png} 
%     \caption{Failure cases. The main failures of our method are: ignore thin structures and filling out empty spaces, demonstrated in these pairs of ground truth and \sald~reconstructions. }
%     \label{fig:failures}
% \end{figure}

\section{Conclusions}

We introduced \sald, a method for learning implicit neural representations from raw data. The method is based on a generalization of the sign agnostic learning idea to include derivative data. We demonstrated that the addition of a sign agnostic derivative term to the loss improves the approximation power of the resulting signed implicit neural network. In particular, showing improvement in the level of details and sharp features of the reconstructions. Furthermore, we identify the favorable minimal surface property of the \sal~and \sald~losses and provide a theoretical justification in 2D. Generalizing this analysis to 3D is marked as interesting future work.

We see two more possible venues for future work: First, it is clear that there is room for further improvement in approximation properties of implicit neural representations. Although the results in D-Faust are already close to the input quality, in ShapeNet we still see a gap between input models and their implicit neural representations; this challenge already exists in overfitting a large collection of diverse shapes in the training stage. Improvement can come from adding expressive power to the neural networks, or further improving the training losses; adding derivatives as done in this paper is one step in that direction but does not solves the problem completely. Combining sign agnostic learning with the recent positional encoding method \citep{tancik2020fourfeat} could also be an interesting future research venue. 
Second, it is interesting to think of applications or settings in which \sald~can improve the current state-of-the-art. Generative 3D modeling is one concrete option, learning geometry with 2D supervision is another. 

%Another interesting future work, \eg, for human modeling, is to think of  certain inductive bias, or even incorporating small number of perfect meshes (\eg, template) so to achieve perfect finger structure, for example. 

\bibliography{sald}
\bibliographystyle{iclr2021_conference}

\appendix
\section{Appendix}
\subsection{Proof of Lemma \ref{lem:sin}}
\label{appendix:proof}
\begin{lemma}\label{lem:sin}
For any pair of unit vectors $\va,\vb$: $\min\set{\norm{\va-\vb}, \norm{\va+\vb}}\geq \abs{\sin \angle(\va,\vb)}$. 
\end{lemma}
\begin{proof}
Let $\va,\vb\in\Real^d$ be arbitrary unit norm vectors. Then,
\begin{align*}
    \min\set{\norm{\va-\vb},\norm{\va+\vb}} &= \brac{\min \set{2+2\ip{\va,\vb}, 2-2\ip{\va,\vb}} }^{1/2} \\
    &= \sqrt{2}\brac{1-\abs{\ip{\va,\vb}}}^{1/2} \\ &= 2 \brac{\frac{1-\abs{\cos \angle(\va,\vb)}}{2}}^{1/2} \\
    & \geq \abs{\sin \angle(\va,\vb)}.
\end{align*}
Where the last inequality can be proved by considering two cases: $\alpha\in [0,\pi/2]$ and $\alpha\in[\pi/2,\pi]$, where we denote $\alpha=\angle(\va,\vb)$. In the first case $\alpha\in[0,\pi/2]$, $\cos \alpha \geq 0$ and in this case $\sqrt{\frac{1-\cos\alpha}{2}} = \abs{\sin\frac{\alpha}{2}}$. The inequality is proved by considering $$2\abs{\sin\frac{\alpha}{2}}-\abs{\sin\alpha} = 2\sin\frac{\alpha}{2}-\sin\alpha = 2\sin\frac{\alpha}{2}(1-\cos\frac{\alpha}{2})\geq 0$$ for $\alpha\in [0,\pi/2]$. For the case $\alpha\in [\pi/2,\pi]$ we have  $\sqrt{\frac{1+\cos\alpha}{2}} = \abs{\cos\frac{\alpha}{2}}$. This case is proved by considering $$2\abs{\cos\frac{\alpha}{2}}-\abs{\sin \alpha}=2\cos\frac{\alpha}{2}-\sin \alpha=2\cos\frac{\alpha}{2}(1-\sin\frac{\alpha}{2})\geq 0$$ for $\alpha\in[\pi/2,\pi]$ 
\end{proof}

\subsection{Implementation Details}
\subsubsection{Data Preparation}\label{appendix:data_prep} 
Given some raw 3D data $\gX$, \sald~loss (See \eqref{e:loss_sald}) is computed on points and corresponding unsigned distance derivatives, $\left\{h\left(\vx\right) \right\}_{\vx\in \gD}$ and $\left\{\nabla_{\vx} h\left( \vx' \right)\right\}_{\vx'\in \gD'}$ (resp.) sampled from some distributions $\gD$ and $\gD'$. In this paper, we set $\gD = \gD_1 \cup \gD_2$, where $\gD_1$ is chosen by uniform sampling points $\left\{\vy\right\}$ from $\gX$ and placing two isotropic Gaussians, $\gN(\vy,\sigma_{1}^2 I)$ and $\gN(\vy,\sigma_{2}^2 I)$ for each $\vy$. The distribution parameter $\sigma_1$ depends on each point $\vy$, set to be as the distance of the \nth{50} closest point to $\vy$, whereas $\sigma_2$ is set to $0.3$ fixed. $\gD_2$ is chosen by projecting $\gD_1$ to $\gS$. The distribution $\gD'$ is set to uniform on $\gX$; note that on $\gX$, $\nabla_\vx h(\vx')$ is a sub-differential which is the convex hull of the two possible normal vectors ($\pm \vn$) at $\vx'$; as the sign-agnostic loss does not differ between the two normal choices, we arbitrarily use one of them in the loss. Computing the unsigned distance to $\gX$ is done using the \textsc{CGAL} library \citep{cgal:eb-20a}. To speed up training, we precomputed for each shape in the dataset, 500K samples of the form $\left\{h\left(\vx\right) \right\}_{\vx\in \gD}$ and $\left\{\nabla_{\vx} h\left( \vx' \right)\right\}_{\vx'\in \gD'}$.

\subsubsection{Gradient computation}
The \sald~loss requires incorporating the term $\nabla_\vx f(\vx;\theta)$ in a differentiable manner. Our computation of $\nabla_\vx f(\vx;\theta)$ is based on \textsc{Automatic Differentiation} \citep{baydin2017automatic} forward mode. Similarly to \cite{gropp2020implicit}, $\nabla_\vx f(\vx;\theta)$ is constructed as a network consists of layers of the form
\[
\nabla_\vx \vy^{\ell+1}=\diag\parr{\sigma'\parr{\mW_{\ell+1}\vy^{\ell}+\vb_{\ell+1}}}\mW_{\ell+1}\nabla_\vx \vy^{\ell}
\]
where $\vy^{\ell}$ denotes the output of the $\ell$ layer in  $f(\vx;\theta)$ and $\vtheta = \left(\mW_{\ell},\vb_{\ell}\right)$ are the learnable parameters.
\subsubsection{Architecture Details}\label{appendix:arch} 
\subsubsection*{VAE Architecture}
Our VAE architecture is based on the one used in \cite{atzmon2019sal}. The encoder $g\left(\mX ; \vtheta_1 \right)$, where $\mX \in \Real^{N\times3}$ is the input point cloud, is composed of  DeepSets \citep{zaheer2017deep} and PointNet \citep{qi2017pointnet} layers. Each layer consists of \begin{align*}
\mathrm{PFC}(d_{\text{in}},d_{\text{out}}):\mX &\mapsto \nu\parr{\mX W + \one b^T }  \\
\mathrm{PL}(d_{\text{in}},2d_{\text{in}}):\mY & \mapsto \brac{\mY, \max{(\mY)} \one}\\
\end{align*}
where $\brac{\cdot,\cdot}$ is the concat operation, $W \in \Real ^{d_{\text{in}}\times d_{\text{out}}}$ and $b\in \Real^{d_{\text{out}}}$ are the layer weights and bias and $\nu\left(\cdot\right)$ is the pointwise non-linear ReLU activation function. Our encoder architecture is: \begin{align*}
&\mathrm{PFC}(3,128) \rightarrow \mathrm{PFC}(128,128) \rightarrow \mathrm{PL}(128,256) \rightarrow\\ 
& \mathrm{PFC}(256,128)  \rightarrow \mathrm{PL}(128,256) \rightarrow \mathrm{PFC}(256,128)  \rightarrow \\
&  \mathrm{PL}(128,256) \rightarrow \mathrm{PFC}(256,128)  \rightarrow \mathrm{PL}(128,256) \rightarrow \\
& \mathrm{PFC}(256,256)  \rightarrow \mathrm{MaxPool} \stackrel{\times 2}{\rightarrow}  \mathrm{FC}(256,256), 
\end{align*}
where $\mathrm{FC}(d_{\text{in}},d_{\text{out}}):\vx \mapsto \nu\parr{W \vx + \vb }$ denotes a fully connected layer. The final two fully connected layers outputs vectors $\vmu\in \Real ^{256} $ and $\veta \in \Real^{256}$ used for parametrization of a multiviariate Gaussian $\gN(\vmu, \diag \exp{\veta})$ used for sampling a latent vector $\vz \in \Real^{256}$. Our encoder architecture is similar to the one used in \cite{mescheder2019occupancy}.

Our decoder $f\left(\brac{\vx,\vz};\vtheta_2\right)$ is a composition of $8$ layers where the first layer is $\mathrm{FC}(256+3,512)$, middle layers are $\mathrm{FC}(512,512)$ and the final layer is $\mathrm{Linear}(512,1)$. Notice that the input for the decoder is $\brac{\vx,\vz}$ where $\vx \in \Real^{3}$ and $\vz$ is the latent vector. In addition, we add a skip connection between the input to the middle fourth layer. We chose the Softplus with $\beta=100$ for the non linear activation in the $\mathrm{FC}$ layers. 
For regulrization of the latent $\vz$, we add the following term to training loss 
$$
0.001 * \left(\norm{\vmu}_1 + \norm{\veta + \one}_1 \right),
$$ similarly to \cite{atzmon2019sal}.
\subsubsection*{Auto-Decoder Architecture}
We use an auto-decoder architecture, similar to the one suggested in \cite{Park_2019_CVPR}. We defined the latent vector $\vz \in \Real^{256}$. The decoder architecture is the same as the one described above for the VAE. For regulrization of the latent $\vz$, we add the following term to the loss
$$
 0.001*\norm{\vz}_2^2,
$$ similarly to \cite{Park_2019_CVPR}.
\subsection{Training details}\label{appendix:training} 
We trained our networks using the \textsc{Adam} \citep{kingma2014adam} optimizer, setting the batch size to $64$. On each training step the \sald~loss is evaluated on a random draw of $92^2$ points out of the precomputed 500K samples. For the VAE, we set a fixed learning rate of $0.0005$, whereas for the AD we scheduled the learning rate to start from $0.0005$ and decrease by a factor of $0.5$ every $500$ epochs. All models were trained for $3000$ epochs. Training was done on $4$ Nvidia V-100 GPUs, using \textsc{pytorch} deep learning framework \citep{paszke2017automatic}.
\subsection{Figures \ref{fig:L} and \ref{fig:U}} \label{appendix:figs}
For the two dimensional experiments in figures \ref{fig:L} and \ref{fig:U} we have used the same decoder as in the VAE architecture with the only difference that the first layer is  $\mathrm{FC}(2,512)$ (no concatenation of a latent vector to the 2D input). We optimized using the \textsc{Adam} \citep{kingma2014adam} optimizer, for $5000$ epochs. The parameter $\lambda$ in the \sald~loss was set to $0.1$.
\subsection{Evaluation}
\label{appendix:eval}
\paragraph{Evaluation metrics.} We use the following Chamfer distance metrics to measure similarity between shapes:
\begin{align} \label{e:CD}
\dist_{\text{C}}\left(\gX_{1},\gX_{2} \right) & = \frac{1}{2}\left(\dist_{\text{C}}^{\rightarrow}\left(\gX_{1},\gX_{2} \right) + \dist_{\text{C}}^{\rightarrow}\left(\gX_{2},\gX_{1} \right) \right)
\end{align}
where 
\begin{align}\label{e:one_sided_CD}
\dist_{\text{C}}^{\rightarrow}\left(\gX_{1},\gX_{2} \right) & = \frac{1}{\abs{\gX_1}}\sum_{\vx_{1}\in\gX_{1}}\min_{\vx_2\in \gX_{2}}\norm{\vx_1-\vx_2}
\end{align}
and the sets $\gX_{i}$ are either point clouds or triangle soups. In addition, to measure similarity of the normals of triangle soups $\gT_1,\gT_2$, we define: 
\begin{align} \label{e:ND}
\dist_{\text{N}}\left(\gT_{1},\gT_{2} \right) & = \frac{1}{2}\left(\dist_{\text{N}}^{\rightarrow}\left(\gT_{1},\gT_{2} \right) + \dist_{\text{N}}^{\rightarrow}\left(\gT_{2},\gT_{1} \right) \right),
\end{align}
where 
\begin{align}\label{e:one_sided_N}
\dist_{\text{N}}^{\rightarrow}\left(\gT_{1},\gT_{2} \right) & = \frac{1}{\abs{\gT_1}}\sum_{\vx_{1}\in\gT_{1}} \angle(\vn(\vx_1), \vn(\hat{\vx}_1)),
%\argmin_{\vx_2\in \gT_{2}}\norm{\vx_1-\vx_2}
\end{align}
where $\angle(\va,\vb)$ is the positive angle between vectors $\va,\vb\in\Real^3$, $\vn\left(\vx_1\right)$ denotes the face normal of a point $\vx_1$ in triangle soup $\gT_1$, and $\hat{\vx}_1$ is the projection of $\vx_1$ on $\gT_2$. 

Tables \ref{tab:shapenet_test} and \ref{tab:dfaust_train} in the main paper report quantitative evaluation of our method, compared to other baselines. The meshing of the learned implicit representation was done using the \textsc{Marching Cubes} algorithm \citep{lorensen1987marching} on a uniform cubical grid of size $\left[512\right]^3$. Computing the evaluation metrics $\dist_C$ and $\dist_N$ is done on a uniform sample of $30$K points from the meshed surface.
% \subsection{More results}
% We provide additional qualitative results for our method from the experiments in section \ref{s:exps} in the main paper. Figure \ref{fig:latent_dfaust}  shows results from our VAE experiment on the D-Faust dataset. The leftmost and rightmost columns are reconstructions obtained by a single forward pass on test (unseen) objects. The middle columns were produced by linear interpolating the leftmost and rightmost latent representations. Notice the ability of \sald~to generate novel mixed faces, body parts and poses. In addition, figure \ref{fig:latent_shapenet} shows results from our Auto-Decoder experiment on the ShapeNet datasets. Similarly to the above, leftmost and rightmost columns are reconstructions of test objects and middle columns show interpolated latent representation reconstructions. 

% \begin{figure}[t]
%     \centering
%     \includegraphics[width=1.0\columnwidth]{figures/supp/all_dfuast_1.png}
%     \caption{Latent interpolation between unseen humans on the D-Faust \cite{bogo2017dynamic} dataset.}
%     \label{fig:latent_dfaust}
% \end{figure}

% \begin{figure}[t]
%     \centering
%     \includegraphics[width=1.0\columnwidth]{figures/supp/allshapebet-1.png}
%     \caption{Latent interpolation between unseen objects from the ShapeNet \cite{chang2015shapenet} dataset.}
%     \label{fig:latent_shapenet}
% \end{figure}
\end{document}